\documentclass{article}
\pdfoutput=1  
\usepackage{style/unites}
\usepackage{XCharter}
\usepackage[scaled=1.1]{zlmtt} 

\usepackage[utf8]{inputenc}
\usepackage[T1]{fontenc}
\usepackage{microtype}

\usepackage{amsmath}
\usepackage{amssymb}
\usepackage{amsfonts}
\usepackage{amsthm}
\usepackage{mathtools}
\usepackage{mathrsfs}
\usepackage{physics}
\usepackage{braket}
\usepackage{slashed}
\usepackage{nicefrac}
\usepackage{textcomp}
\usepackage{dsfont}
\usepackage{bbm}
\usepackage{bm}

\usepackage{graphicx}
\usepackage{subcaption}
\usepackage[export]{adjustbox}
\usepackage{float}
\usepackage{booktabs}
\usepackage{dcolumn}
\newcolumntype{d}[1]{D{.}{.}{#1}}
\usepackage{bigstrut, tabularx, multirow, makecell, diagbox}
\usepackage{colortbl}
\usepackage{tabularray}
\UseTblrLibrary{booktabs}
\usepackage{threeparttable}
\usepackage{tablefootnote}
\usepackage{fontawesome5}

\usepackage{placeins}
\usepackage{caption}
\usepackage{footnote}
\usepackage{enumitem}
\usepackage{multicol}
\usepackage{xspace}
\usepackage{titletoc}
\usepackage{titlesec}
\usepackage[bottom]{footmisc}
\usepackage{setspace}

\usepackage{wrapfig}
\usepackage{tikz}
\usepackage{quantikz}
\usepackage{dashbox}
\usepackage{mdframed}
\usepackage{marvosym}
\usepackage{pifont}
\usepackage{CJK}
\usepackage{url}

\usepackage[table,x11names]{xcolor}
\usepackage[most]{tcolorbox}
\tcbuselibrary{breakable}
\usetikzlibrary{decorations.pathreplacing, fit}

\definecolor{primaryblue}{HTML}{0066CC}
\definecolor{accentcyan}{HTML}{00D4AA}
\definecolor{warmorange}{HTML}{FF6B35}
\definecolor{deepgray}{HTML}{2C3E50}
\definecolor{lightgray}{HTML}{F8F9FA}
\definecolor{gradientstart}{HTML}{667eea}
\definecolor{gradientend}{HTML}{764ba2}

\definecolor{citecolor}{HTML}{0071bc}
\definecolor{citeblue}{RGB}{0, 113, 188}
\definecolor{linkcolor}{HTML}{9A4D92}
\definecolor{firebrick}{rgb}{0.698,0.133,0.133}

\definecolor{paleviolet}{HTML}{E1EEFC}
\definecolor{CarolinaUltraLight}{HTML}{E7F4FC}
\definecolor{lightgrey}{RGB}{247, 247, 247}
\definecolor{shadecolor}{HTML}{EFEFEF}
\definecolor{lightyellow}{rgb}{1.0, 0.95, 0.7}
\definecolor{lightblue}{rgb}{0.90, 0.95, 1.0}
\definecolor{light-gray}{gray}{0.95}

\definecolor{darkgrey}{rgb}{0.5, 0.5, 0.5}
\definecolor{darkgreen}{rgb}{0, 0.5, 0}
\definecolor{mydarkblue}{rgb}{0,0.08,0.45}
\definecolor{mydarkblue2}{rgb}{0.133, 0.133, 0.698}
\definecolor{echodrk}{HTML}{0099cc}
\definecolor{mymauve}{rgb}{0.58,0,0.82}
\definecolor{midnightblue}{rgb}{0.1,0.1,0.44}
\definecolor{oxfordblue}{rgb}{0.0,0.13,0.28}
\definecolor{prussianblue}{rgb}{0.0,0.19,0.33}
\definecolor{coolteal}{rgb}{0, 0.45, 0.45}
\definecolor{olive}{rgb}{0.1, 0.3, 0}
\definecolor{mypurple}{rgb}{0.5,0,0.5}
\definecolor{almond}{rgb}{0.94, 0.87, 0.8}

\definecolor{blue_ampEncoding}{HTML}{DAE8FC}
\definecolor{green_encoder}{HTML}{D5E8D4}
\definecolor{purple_decoder}{HTML}{E1D5E7}
\definecolor{yellow_measure}{HTML}{FFF2CC}
\definecolor{gray_block}{HTML}{F5F5F5}
\definecolor{pink_dru}{HTML}{FAD9D5}
\definecolor{orange_v}{HTML}{FAD7AC}

\definecolor{colorA}{rgb}{1,0,0}
\definecolor{colorB}{rgb}{0,0.3,1}
\definecolor{colorC}{rgb}{0.9,0.8,0.2}
\definecolor{colorD}{rgb}{0,0.65,0}
\definecolor{lesslightgray}{rgb}{0.5,0.5,0.5}
\definecolor{fundamental}{RGB}{55, 110, 111}
\definecolor{Gred}{RGB}{219, 50, 54}
\definecolor{ToCgreen}{RGB}{0, 128, 0}
\definecolor{Sepia}{RGB}{112, 66, 20}
\definecolor{Dblue}{rgb}{0,0.08,0.45}
\definecolor{Blue}{rgb}{0, 0, 0.8}
\definecolor{blue}{rgb}{0,0,1}
\definecolor{UNCblue!10}{rgb}{0.84,0.91,0.98}
\definecolor{RowAlt}{rgb}{0.98,0.98,0.99}

\definecolor{CarolinaBlue}{HTML}{7BAFD4}        
\definecolor{CarolinaLightBlue}{HTML}{B3D4E5}   
\definecolor{CarolinaUltraLight}{HTML}{E8F4F8}  
\definecolor{CarolinaText}{HTML}{1C2B33}        

\usepackage[pagebackref=true,breaklinks=true,colorlinks,hyperfootnotes=false]{hyperref}
\hypersetup{
  colorlinks,
  citecolor=citeblue,
  linkcolor=firebrick,
  urlcolor=firebrick
}
\usepackage[nameinlink,capitalize,noabbrev]{cleveref}

\titlespacing\section{0pt}{4pt plus 4pt minus 2pt}{-2pt plus 2pt minus 2pt}
\titlespacing\subsection{0pt}{2pt plus 4pt minus 2pt}{-2pt plus 2pt minus 2pt}
\titlespacing\subsubsection{0pt}{2pt plus 4pt minus 2pt}{-2pt plus 2pt minus 2pt}

\makeatletter
\def\th@remark{%
  \thm@headfont{\bfseries}%
  \normalfont 
  \thm@preskip\topsep \divide\thm@preskip\tw@
  \thm@postskip\thm@preskip
}
\makeatother

\theoremstyle{definition}

\tcolorboxenvironment{theorem}{
  breakable,
  colback=black!10,
  colframe=white,
  width=\linewidth, 
  enlarge left by=0pt,
  enlarge right by=0pt,
  boxsep=5pt,
  boxrule=0pt,
  left=0pt,right=0pt,top=0pt,bottom=0pt,
  arc=8pt,
  before skip=\topsep,
  after skip=\topsep
}

\tcolorboxenvironment{lemma}{
  breakable,
  colback=black!10,
  colframe=white,
  width=\linewidth,
  enlarge left by=0pt,
  enlarge right by=0pt,
  boxsep=5pt,
  boxrule=0pt,
  left=0pt,right=0pt,top=0pt,bottom=0pt,
  arc=8pt,
  before skip=\topsep,
  after skip=\topsep
}

\tcolorboxenvironment{corollary}{
  breakable,
  colback=black!10,
  colframe=white,
  width=\linewidth,
  enlarge left by=0pt,
  enlarge right by=0pt,
  boxsep=5pt,
  boxrule=0pt,
  left=0pt,right=0pt,top=0pt,bottom=0pt,
  arc=8pt,
  before skip=\topsep,
  after skip=\topsep
}

\tcolorboxenvironment{proposition}{
  breakable,
  colback=black!10,
  colframe=white,
  width=\linewidth,
  enlarge left by=0pt,
  enlarge right by=0pt,
  boxsep=5pt,
  boxrule=0pt,
  left=0pt,right=0pt,top=0pt,bottom=0pt,
  arc=8pt,
  before skip=\topsep,
  after skip=\topsep
}

\tcolorboxenvironment{definition}{
  breakable,
  colback=black!10,
  colframe=white,
  width=\linewidth,
  enlarge left by=0pt,
  enlarge right by=0pt,
  boxsep=5pt,
  boxrule=0pt,
  left=0pt,right=0pt,top=0pt,bottom=0pt,
  arc=8pt,
  before skip=\topsep,
  after skip=\topsep
}

\tcolorboxenvironment{assumption}{
  breakable,
  colback=black!10,
  colframe=white,
  width=\linewidth,
  enlarge left by=0pt,
  enlarge right by=0pt,
  boxsep=5pt,
  boxrule=0pt,
  left=0pt,right=0pt,top=0pt,bottom=0pt,
  arc=8pt,
  before skip=\topsep,
  after skip=\topsep
}

\tcolorboxenvironment{claim}{
  breakable,
  colback=black!10,
  colframe=white,
  width=\linewidth,
  enlarge left by=0pt,
  enlarge right by=0pt,
  boxsep=5pt,
  boxrule=0pt,
  left=0pt,right=0pt,top=0pt,bottom=0pt,
  arc=8pt,
  before skip=\topsep,
  after skip=\topsep
}

\tcolorboxenvironment{problem}{
  breakable,
  colback=black!10,
  colframe=white,
  width=\linewidth,
  enlarge left by=0pt,
  enlarge right by=0pt,
  boxsep=5pt,
  boxrule=0pt,
  left=0pt,right=0pt,top=0pt,bottom=0pt,
  arc=8pt,
  before skip=\topsep,
  after skip=\topsep
}

\tcolorboxenvironment{question}{
  breakable,
  colback=black!10,
  colframe=white,
  width=\linewidth,
  enlarge left by=0pt,
  enlarge right by=0pt,
  boxsep=5pt,
  boxrule=0pt,
  left=0pt,right=0pt,top=0pt,bottom=0pt,
  arc=8pt,
  before skip=\topsep,
  after skip=\topsep
}

\newtcolorbox{titleblock}{
  enhanced,
  frame hidden,
  colback=CarolinaUltraLight,
  colframe=CarolinaUltraLight,
  boxrule=0pt,
  arc=10pt,
  left=14pt,
  right=14pt,
  top=14pt,
  bottom=14pt,
  width=\linewidth,
  before skip=12pt plus 4pt,
  after skip=12pt plus 4pt,
  grow to left by=1.5pt,
  grow to right by=1.5pt,
  before upper={
    \setlength{\parindent}{0cm}
    \setlength{\parskip}{0.5cm}
  }
}

\crefname{theorem}{Theorem}{Theorems}
\crefname{proposition}{Proposition}{Propositions}
\crefname{lemma}{Lemma}{Lemmas}
\crefname{corollary}{Corollary}{Corollaries}
\crefname{definition}{Definition}{Definitions}
\crefname{assumption}{Assumption}{Assumptions}
\crefname{remark}{Remark}{Remarks}
\crefname{problem}{Problem}{Problems}
\crefname{property}{Property}{property}
\crefname{question}{Question}{Questions}

\numberwithin{equation}{section}
\numberwithin{theorem}{section}
\numberwithin{proposition}{section}
\numberwithin{definition}{section}
\numberwithin{lemma}{section}
\numberwithin{assumption}{section}
\numberwithin{remark}{section}

\def\1{\bm{1}}

\makeatletter
\let\save@mathaccent\mathaccent
\newcommand*\if@single[3]{%
    \setbox0\hbox{${\mathaccent"0362{#1}}^H$}%
    \setbox2\hbox{${\mathaccent"0362{\kern0pt#1}}^H$}%
    \ifdim\ht0=\ht2 #3\else #2\fi
}
\newcommand*\rel@kern[1]{\kern#1\dimexpr\macc@kerna}
\newcommand*\widebar[1]{\@ifnextchar^{{\wide@bar{#1}{0}}}{\wide@bar{#1}{1}}}
\newcommand*\wide@bar[2]{\if@single{#1}{\wide@bar@{#1}{#2}{1}}{\wide@bar@{#1}{#2}{2}}}
\newcommand*\wide@bar@[3]{%
    \begingroup
    \def\mathaccent##1##2{%
        \let\mathaccent\save@mathaccent
        \if#32 \let\macc@nucleus\first@char \fi
        \setbox\z@\hbox{$\macc@style{\macc@nucleus}_{}$}%
        \setbox\tw@\hbox{$\macc@style{\macc@nucleus}{}_{}$}%
        \dimen@\wd\tw@
        \advance\dimen@-\wd\z@
        \divide\dimen@ 3
        \@tempdima\wd\tw@
        \advance\@tempdima-\scriptspace
        \divide\@tempdima 10
        \advance\dimen@-\@tempdima
        \ifdim\dimen@>\z@ \dimen@0pt\fi
        \rel@kern{0.6}\kern-\dimen@
        \if#31
        \overline{\rel@kern{-0.6}\kern\dimen@\macc@nucleus\rel@kern{0.4}\kern\dimen@}%
        \advance\dimen@0.4\dimexpr\macc@kerna
        \let\final@kern#2%
        \ifdim\dimen@<\z@ \let\final@kern1\fi
        \if\final@kern1 \kern-\dimen@\fi
        \else
        \overline{\rel@kern{-0.6}\kern\dimen@#1}%
        \fi
    }%
    \macc@depth\@ne
    \let\math@bgroup\@empty \let\math@egroup\macc@set@skewchar
    \mathsurround\z@ \frozen@everymath{\mathgroup\macc@group\relax}%
    \macc@set@skewchar\relax
    \let\mathaccentV\macc@nested@a
    \if#31
    \macc@nested@a\relax111{#1}%
    \else
    \def\gobble@till@marker##1\endmarker{}%
    \futurelet\first@char\gobble@till@marker#1\endmarker
    \ifcat\noexpand\first@char A\else
    \def\first@char{}%
    \fi
    \macc@nested@a\relax111{\first@char}%
    \fi
    \endgroup
    }
\makeatother

\DeclareMathAlphabet{\mathsfit}{\encodingdefault}{\sfdefault}{m}{sl}
\SetMathAlphabet{\mathsfit}{bold}{\encodingdefault}{\sfdefault}{bx}{n}

\let\tilde\widetilde
\let\hat\widehat

\renewcommand{\arraystretch}{1.15}
\newlength{\aaaicolwidth}
\newcommand{\ourmethod}{ReCoGen}

\begin{document}

\makeatletter
\def\blfootnote{\gdef\@thefnmark{}\@footnotetext}
\makeatother

\makeatletter
\pagestyle{fancy}
\fancyhf{}
\renewcommand{\headrulewidth}{1pt}
\chead{\small\bf Represent, Then Generate: Multimodal-Conditioned Time-Series Generation under Irregular Missingness
}
\cfoot{\thepage}
\thispagestyle{fancy}
\makeatother

\makeatletter
\def\icmldate#1{\gdef\@icmldate{#1}}
\icmldate{\today}
\makeatother

\makeatletter
\fancypagestyle{fancytitlepage}{
  \fancyhead{}
  \lhead{\includegraphics[height=0.8cm]{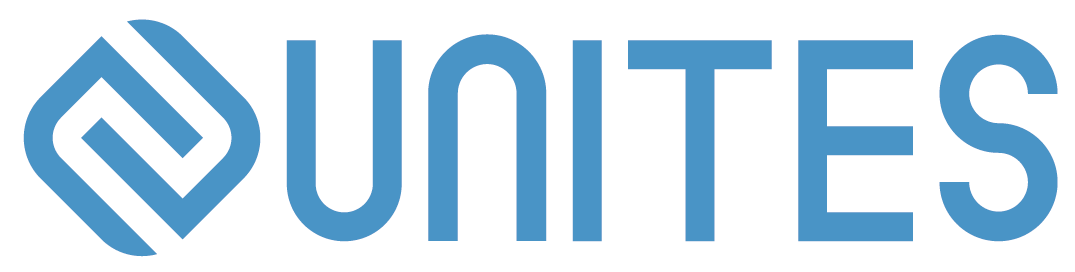}}
  \rhead{\it \@icmldate}
  \cfoot{}
}
\makeatother

\thispagestyle{fancytitlepage}

\vspace*{0.5em}

\noindent
\begin{titleblock}
    {\setlength{\parskip}{0cm}
     \raggedright
     {\setstretch{1.2}
      \LARGE\bfseries
      Represent, Then Generate: Multimodal-Conditioned Time-Series Generation under Irregular Missingness
      \par}
    }
    \vskip 0.2cm
    
\begin{icmlauthorlist}
\mbox{Haochen Zhang$^{\,1\,}$},
\mbox{Jiaheng Guo$^{\,1\,}$},
\mbox{Yu-Chao Huang$^{\,1\,}$},
\mbox{Nicholas Konz$^{\,1\,}$}
and \mbox{Tianlong Chen$^{\,1\,\textrm{\Letter}}$}
\end{icmlauthorlist}

$^{1\,}$UNITES Lab, University of North Carolina at Chapel Hill

\{haochenz, morris, nick124, tianlong\}@cs.unc.edu, jiaheng@unc.edu

$^{\textrm{\Letter}}$ Corresponding Author

    \vskip 0.2cm
    
    Continuous physiological time series underpin modern clinical monitoring, yet many of the most informative signals are invasive, expensive, or simply unavailable for a given patient.
Conditional generation offers a remedy: an absent signal can be synthesized from co-recorded signals and routine clinical variables. Existing generators, however, are built around a single conditioning modality and degrade when forced to handle the heterogeneous, irregularly missing mix of time-variant signals and static covariates seen in practice.
We propose \ourmethod{} (\textbf{Re}present \textbf{Co}nditions, then \textbf{Gen}erate), a two-stage framework that decouples multimodal condition representation from target generation. Stage~I trains one masked autoencoder per modality, distilling each time-variant condition into a compact and missingness-tolerant token sequence. 
Stage~II trains a flow-matching generator that fuses these tokens with static conditions to synthesize the target signal. Across three physiological benchmarks, including continuous glucose monitoring on AI-READI and arterial blood pressure generation on MIMIC-III and MIMIC-IV, \ourmethod{} attains the best downstream utility on all sixteen (dataset, task, metric) settings, surpassing six representative conditional generators; on thirteen of them its utility also reaches or exceeds the utility measured on the real signal, a reference we read as an approximate anchor rather than a ceiling. Ablations trace the gains to the conditioning path: learnable cross-attention over the frozen per-modality encoders, and a dual token-plus-AdaLN route for the static conditions. \ourmethod{} thus turns routinely collected signals into informative surrogates for invasive or unavailable ones, a step toward less invasive, lower-cost continuous clinical monitoring.

\end{titleblock}

\blfootnote{%
$^{\textrm{\Letter}}$ Corresponding authors: \{tianlong\}@cs.unc.edu
\\[2.5em]
\ifcsname @icmlpreprint\endcsname
  \textit{\csname @icmlpreprint\endcsname}%
\fi
}


\section{Introduction}
\label{sec:intro}
Continuous physiological time series have become a foundation of modern medicine \citep{topol2019high,dunn2018wearables}: bedside monitors and consumer wearables now collect cardiac, respiratory, and metabolic signals that trace a patient's health state and support diagnosis, risk stratification, and the early warning of adverse events. Intensive-care vital-sign streams support prediction of in-hospital mortality~\citep{harutyunyan2019multitask,alghatani2021los}, early sepsis onset~\citep{nemati2018sepsis}, and impending circulatory failure~\citep{hyland2020circulatory}; wearable continuous glucose monitoring enables more precise screening for diabetes \citep{hall2018glucotypes,lu2025cgmformer}.

However, many informative signals are difficult and costly to obtain. Continuous arterial blood pressure (ABP) acquisition is invasive and carries procedural risk, so it is used for only a minority of patients \citep{nuttall2016surgical}; and signals that are easy to collect often arrive with irregular missingness, which materially complicates downstream use \citep{che2018recurrent}. Conditional time series generation is a remedy: synthesize the hard-to-measure target from available co-recorded modalities, as in cross-modal biosignal synthesis~\citep{sarkar2021cardiogan} and generative augmentation of physiological recordings~\citep{li2022ttscgan}, so that monitoring and analysis extend to settings where the target is unavailable or hard to access \citep{lai2025diffusets}.

\begin{wrapfigure}{r}{\aaaicolwidth}
\centering
    \includegraphics[width=\linewidth]{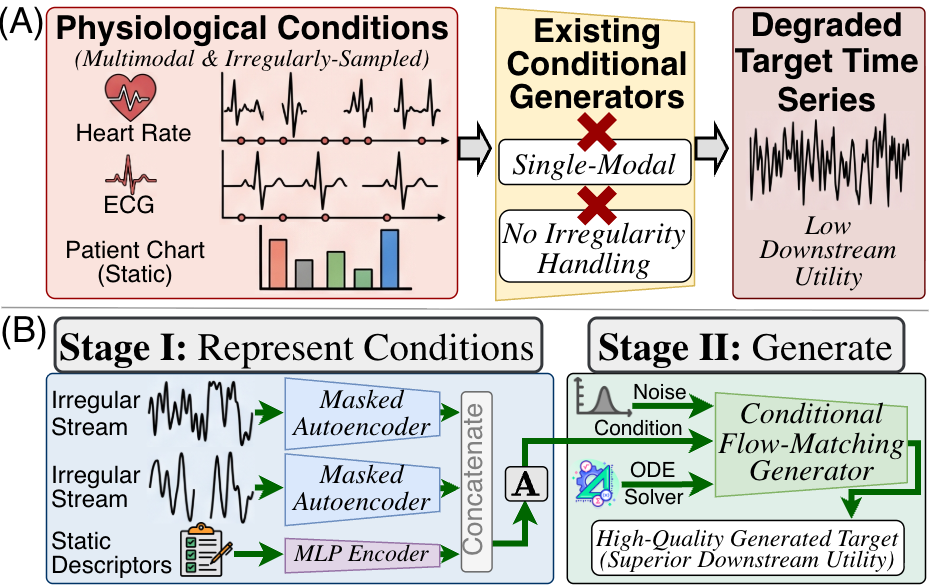}
    \caption{Motivation and overview. (A) Existing conditional generators ingest multimodal physiological conditions directly and fail on heterogeneous modalities with irregular missingness. (B) \ourmethod{} first represents each condition modality with missingness-aware encoders, then generates the target with a conditional flow-matching model.}
    \label{fig:teaser}
\end{wrapfigure}


Conditional time series generation methods have been studied for handling a wide range of conditioning signals \citep{narasimhan2024timeweaver}. Label-conditioned methods synthesize time series from discrete class labels \citep{li2022ttscgan,lee2023timevqvae}. Attribute-conditioned methods condition on structured metadata~\citep{jing2024tedit,shankar2025wavestitch}. Text-conditioned models convert the conditions into natural language description to enable flexibly controllable generation \citep{gu2025verbalts,li2025bridge,lai2025diffusets}. Imputation methods infill the missing part conditioned on the observed proportion of a multivariate time series \citep{tashiro2021csdi,yuan2024diffusionts,naiman2024imagentime}. However, these methods are mainly built for a single conditioning modality, whereas clinical conditioning signals are inevitably multimodal: the target must be generated jointly from several co-recorded time series together with static tabular and categorical clinical variables. Even methods designed for heterogeneous covariates \citep{narasimhan2024timeweaver} do not address irregularly sampled covariates or modality missingness, which characterize wearable and ICU data~\citep{che2018recurrent}. Representing and merging conditions across modalities to build a robust conditional generator remains an open problem.

The most straightforward way to build such a system from existing methods is to fold the conditions from multiple modalities directly into the generator: treating every co-recorded time series as observed context, as in imputation methods \citep{tashiro2021csdi,yuan2024diffusionts,naiman2024imagentime}, and handling the static conditions with attribute-conditioned methods \citep{narasimhan2024timeweaver}. Multimodal physiological conditions are precisely where this shortcut breaks down (Figure~\ref{fig:teaser}A), for two reasons. First, co-recorded time series arise from distinct physiological processes with their own dynamics and clinical meaning~\citep{baltrusaitis2018multimodal,acosta2022multimodal}, so an in-painting mechanism that treats them as homogeneous context cannot represent these modality-specific semantics. Second, each modality has its own irregular sampling grid and missingness pattern, making it hard to learn representations of the time-variant conditions while performing generation. Indeed, handing the raw, partially observed streams straight to the generator measurably degrades the synthesized target on the harder downstream tasks (Figure~\ref{fig:ablation-tscond}). This leads us to a simple hypothesis: properly representing a condition deserves as much attention as generating the target.

We validate this hypothesis and propose \ourmethod{}: a two-stage framework for multimodal-conditioned physiological time-series generation that decouples condition representation from target generation (Figure~\ref{fig:teaser}B). In Stage~I we learn one masked autoencoder per modality \citep{he2022masked}, so each encoder imputes from context and distills an irregular, time-variant modality into a compact, robust token sequence. In Stage~II we freeze these encoders and train a flow-matching model \citep{lipman2023flow,liu2023rectified} that synthesizes the target from both static and time-variant conditions, where we find the conditioning-path design has a substantial effect on downstream utility. Our contributions are as follows:
\begin{itemize}
\item We show that the current methods for conditional time-series
generation perform poorly under multimodal physiological conditioning; and we trace the failure to the absence of a dedicated, missingness-aware condition representation.
\item We propose \ourmethod{}, a two-stage framework that decouples condition
representation from generation: per-modality masked autoencoders distill each
irregular modality into missingness-tolerant tokens; and the conditional generator can merge conditioning features in a separate training stage.
\item We conduct comprehensive experiments across three physiological datasets, showing that \ourmethod{} outperforms six representative conditional generators in downstream task utility; through a systematic ablation study, we explain the intuition behind the design of \ourmethod{}.

\end{itemize}


\section{Method}
\label{sec:method}

\subsection{Problem Setup}

We study the generation of a physiological time series from the other signals and
clinical variables recorded for the same subject. The target $X$ is valued in
$\mathbb{R}^{V\times T}$ ($V$ channels of length $T$). The conditions come in two
forms: $M$ co-recorded time-variant modalities, the $m$-th valued in
$\mathbb{R}^{T_m}$ and possibly irregularly sampled, each paired with a binary
observed mask $o^{(m)}\in\{0,1\}^{T_m}$ ($1$ at observed steps, $0$ at missing
ones); and a static descriptor $s$ in an abstract space $\mathcal{S}$ ---
questionnaire responses, laboratory tests, or medical history, varying across
cohorts. Writing $c=(c^{(1)},o^{(1)},\dots,c^{(M)},o^{(M)},s)$ for a realization of
the joint condition $C$, the goal is to learn
$\hat{P}_{\theta}(X \mid C) \approx P(X\mid C)$ from i.i.d.\ samples
$\{(x_i,c_i)\}_{i=1}^{N}$ of the unknown joint distribution $P$.

We take a data-augmentation view of generation quality: generated signals should be
as useful as real ones for a downstream clinical predictor. Fix a label space
$\mathcal{Y}$, a hypothesis class $\mathcal{H}$ of predictors
$h:\mathbb{R}^{V\times T}\!\to\mathcal{Y}$ and a loss $L$; write
$R_{Q}(h)=\mathbb{E}_{(x,\ell)\sim Q}[L(h(x),\ell)]$ for the risk under a
distribution $Q$ over signal--label pairs, and let
$h^{\star}=\arg\min_{h\in\mathcal{H}} R_{P}(h)$ be the predictor a practitioner
would train on real data. Our generator induces a synthetic distribution
$\hat{P}_{\theta}$, under which a signal is generated from the real conditions and
paired with the same label. Generation serves augmentation well when the
generalization gap
\begin{equation}
\Delta(\theta)=\mathbb{E}_{(c,\ell)}
\mathbb{E}_{\hat{x}\sim \hat{P}_{\theta}(\cdot\mid c)}\big[L(h^{\star}(\hat{x}),\ell)\big]
-R_{P}(h^{\star})
\label{eq:gap}
\end{equation}
is small in magnitude; we estimate both risks on held-out subjects and report them as
AUROC and AUPRC, for which a lower risk is a higher value. Three qualifications fix
how $\Delta(\theta)$ should be read: the quantity to control is $|\Delta(\theta)|$,
since the gap is signed and a large negative gap is not better than a zero gap; it is
informative only when $R_{P}(h^{\star})$ is itself non-trivial, as otherwise both
risks sit near the floor and their agreement certifies nothing; and it is necessary
but not sufficient for fidelity, since a generator preserving only the label-relevant
component of $x$ also attains $\Delta(\theta)\approx 0$. In particular
$\Delta(\theta)<0$ calls for scrutiny rather than celebration: it is the signature
one would expect if the generator re-expressed the conditions in a form the predictor
reads more easily than the measured waveform, a point we return to with the main
results.

\subsection{Overview of \ourmethod{}}

\begin{figure}[!t]
\centering
\includegraphics[]{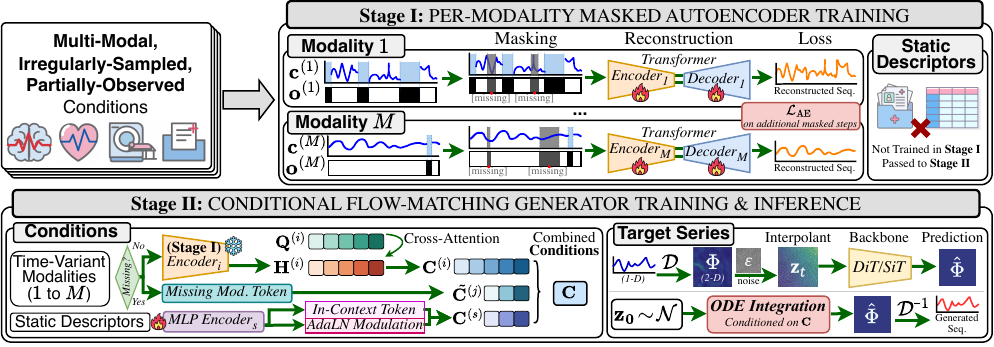}
\caption{Overview of \ourmethod{}. Stage~I trains one masked autoencoder per
time-variant conditioning modality; Stage~II freezes them and trains a conditional
flow-matching generator fusing the time-variant and static conditions to synthesize
the target series.}
\label{fig:overview}
\end{figure}

\ourmethod{} generates a target from its conditions in two decoupled stages
(Figure~\ref{fig:overview}). For each irregular time-variant modality, Stage~I
learns a robust encoder by training a lightweight Transformer encoder--decoder over
the full-length event stream under a masked-reconstruction objective that forces the
encoder to impute from context. Stage~II freezes those encoders and trains a
conditional flow-matching generator that combines the time-variant and the static
conditions to synthesize the target series. Both stages operate in the observation
domain: the generator represents a 1-D series as a 2-D image through an invertible
delay embedding~\citep{naiman2024imagentime} and models it with a vision transformer.

\ourmethod{} rests on two core designs. The first is the decoupled training just
outlined: the feature extractor for the time-variant conditions is trained on its own
masked-reconstruction objective and then frozen, so the generator is optimized on top
of fixed conditioning features instead of having to learn how to encode irregular,
partially observed series jointly with generation. The second is the
conditioning path. Each time-variant modality is summarized by a bank of learnable
per-modality query tokens that cross-attend over its frozen autoencoder latents,
giving a compact set of in-context condition tokens; the static descriptor enters
through two complementary routes, as an additional in-context token and as an
adaptive layer-norm (AdaLN) modulation. This path accounts for most of the
generation quality.

\subsection{Stage~I: Masked Autoencoder Training}
We learn the representation of each time-variant conditioning modality with a masked autoencoder. Take one modality and write its series as $u=(u_1,\dots,u_L)$ with observed mask $o\in\{0,1\}^{L}$ (a generic $c^{(m)}$ and $o^{(m)}$), where $o_t=1$
marks a genuinely observed step and $o_t=0$ a step that is missing. The
observed steps $\mathcal{O}=\{t:o_t=1\}$ already leave a modality-dependent fraction
of the series missing; on top of this natural missingness we sample a random subset
$\mathcal{M}\subseteq\mathcal{O}$ with mask ratio $\rho$ and hold it out for
reconstruction. Every naturally missing or held-out step is replaced by a single
shared learnable \texttt{[missing]} token, so the encoder always sees a
regular-length sequence. Following
masked autoencoding~\citep{he2022masked}, the loss is the mean squared error on the
held-out steps only,
\begin{equation}
\mathcal{L}_{\mathrm{AE}}
=\frac{1}{|\mathcal{M}|}\sum_{t\in\mathcal{M}}\big(\hat{u}_t-u_t\big)^2 .
\label{eq:ae}
\end{equation}
Reconstructing unseen points forces the encoder to infer from context rather than copy its input, which keeps it robust when a modality is sparsely observed. We train one autoencoder $\mathrm{AE}_m$ per modality $m$, freeze it for Stage~II, and use $\rho{=}0.3$ by default.

\subsection{Stage~II: Conditional Flow-Matching Training}

Stage~II trains a conditional generator on top of the frozen Stage~I
encoders, which now act as fixed feature extractors; on the conditioning side only
the injection modules remain trainable, which also stabilizes training. A time series
$x\in\mathbb{R}^{V\times T}$ is mapped to a square image
$\Phi=\mathcal{D}(x)\in\mathbb{R}^{V\times e\times e}$ by a delay
embedding~\citep{naiman2024imagentime}, whose $i$-th column is the non-overlapping
length-$e$ window $x[\,ie:(i{+}1)e\,]$ (embedding dimension $e$), padded to an
$e\times e$ square. The map $\mathcal{D}$ is invertible on the valid region, so a
generated image is read back to a series by $\mathcal{D}^{-1}$, and a binary mask
$\Phi_{\text{mask}}$ marks the non-padded pixels, on which alone the flow-matching
loss is computed.

We investigate the conditioning-path design for multimodal conditional time-series
generation. For modality $m$, the frozen encoder yields a latent sequence
$H^{(m)}=\mathrm{AE}_m^{\text{enc}}(c^{(m)},o^{(m)})$, which a bank of $n$ learnable
per-modality query tokens $Q^{(m)}$ summarizes by cross-attention,
$A^{(m)}=\mathrm{CrossAttn}(Q^{(m)},H^{(m)},H^{(m)})\in\mathbb{R}^{n\times d}$, giving
$M\!\cdot\!n$ time-variant condition tokens; a missing modality falls back to a
per-modality learnable missing-token sequence $\tilde{A}^{(m)}\in\mathbb{R}^{n\times d}$.
These tokens enter the generator in context. How they are injected is a key design
choice: learnable cross-attention pooling outperforms plain in-context tokens,
modulation-only injection, and an in-painting alternative, especially on the harder
ABP tasks (Figure~\ref{fig:ablation-tscond}). The static descriptor is likewise
encoded (with its missingness made visible) into one or two tokens, depending on the
cohort's static schema, and injected
through both routes, which together outperform either alone
(Table~\ref{tab:ablation-static}).

The denoiser backbone $f_\theta$ is a DiT/SiT-style vision
transformer~\citep{peebles2023scalable}, trained with the $x$-prediction $v$-loss
flow-matching formulation~\citep{li2026jit}. With noise
$\epsilon\sim\mathcal{N}(0,\sigma^2 I)$ and $t\in(0,1)$ from a logit-normal schedule,
we form the interpolant $z_t=t\,\Phi+(1-t)\,\epsilon$ with target velocity
$v^\star=\Phi-\epsilon$; the network predicts the clean signal $\hat\Phi=f_\theta(z_t,t,c)$,
which gives $\hat v=(\hat\Phi-z_t)/(1-t)$, and we minimize the mask-normalized error
$\mathbb{E}_{\Phi,\epsilon,t}\big[\lVert \Phi_{\text{mask}}\odot(\hat v-v^\star)\rVert_2^2/\lVert \Phi_{\text{mask}}\rVert_1\big]$.
At inference we integrate the ODE $dz_t/dt=\hat v$ from noise ($t{=}0$) to data
($t{=}1$) with a fixed number of Euler or second-order Heun steps (the last always
Euler), encoding the conditions once and mapping the final image back to a series
with $\mathcal{D}^{-1}$.

\section{Experiment}

\subsection{Experiment Setup}
\label{sec:exp-setup}

\paragraph{Datasets and tasks}
We evaluate on three tasks spanning two domains: one wearable/glycemic cohort and two
critical-care databases. All windows are 24 hours at 5-minute resolution ($T{=}288$).
\textbf{(i) AI-READI}~\citep{aireadi2024}, a multimodal type-2-diabetes cohort: we
generate a subject's CGM trace from four co-recorded wearable modalities (heart rate,
calorie expenditure, physical activity, respiratory rate) and static tabular clinical
features. \textbf{(ii) MIMIC-III}~\citep{johnson2016mimic}: we generate the mean ABP
trace from three co-recorded vitals (heart rate, respiratory rate,
$\mathrm{SpO}_2$) and a static clinical vector, all taken from the Waveform Database
Matched Subset numerics. \textbf{(iii) MIMIC-IV}~\citep{johnson2023mimic} poses the
same task on a distinct cohort, but from bedside charted measurements rather than
waveforms, giving a dual-resolution setup with the sparsely charted vitals on an
hourly grid ($T_{\mathrm{cond}}{=}24$).

\paragraph{Baselines}
We compare against six representative conditional generators, grouped by the
conditioning interface their architecture exposes; all generate the same target as
\ourmethod{}.
\emph{Signal-conditioned:} Diffusion-TS~\cite{yuan2024diffusionts} and
ImagenTime~\cite{naiman2024imagentime} stack the target and the co-recorded series
into one multivariate series and in-paint the target, supervising the target channel
only; neither admits a label or static features, so both consume the conditioning
series alone.
\emph{Attribute-conditioned:} TimeWeaver~\cite{narasimhan2024timeweaver} fuses the
conditioning series as time-varying metadata into a CSDI-style denoiser, and
WaveStitch~\cite{shankar2025wavestitch} pins the co-recorded channels as RePaint
anchors on an SSSD-S4 backbone; neither has a path for the tabular static vector. We
take both from ConTSG-Bench~\citep{contsgbench2026}, a public benchmark whose ports
condition on discrete attributes alone, and add back each paper's own mechanism:
TimeWeaver's continuous metadata path and WaveStitch's anchor conditioning.
\emph{Text-conditioned:} VerbalTS~\cite{gu2025verbalts} and Bridge~\cite{li2025bridge}
accept text only, so we verbalize the complete conditioning set and embed it with a
frozen LongCLIP text encoder; Bridge additionally draws a same-label real target
window as its prototype. These two therefore receive our full conditioning set.
Per-method settings are in the supplement.

\paragraph{Fairness of the comparison}
Everything outside that interface is held fixed: all baselines reuse our dataset
builder, split, windowing and valid-window filter verbatim, generate on the same
evaluation windows, are scored by the same probe, and train for 1000 epochs at
learning rate $10^{-4}$ with the optimizer of their reference implementation. Each
receives the largest subset of the conditioning set its architecture admits ---
complete for VerbalTS and Bridge, and Bridge in fact receives strictly more, since
its prototype is a real target window. That the other four
cannot ingest heterogeneous static conditions is the limitation this work targets,
not an artifact of our setup; Figure~\ref{fig:ablation-tscond} isolates it with an
in-painting variant of \ourmethod{} that changes only how the conditioning series
enter.

\begin{table}[!t]
\centering
\newcommand{\hd}[2]{\makecell{#1\\#2}}
\begin{tabular}{ll cccccc c c}
\toprule
& & \multicolumn{6}{c}{Generative baselines} & \multirow{2}{*}{\textbf{\ourmethod{}}} & Ref. \\
\cmidrule(lr){3-8}\cmidrule(lr){10-10}
Task & Metric
& \hd{Diffusion}{-TS} & \hd{Imagen}{Time} & \hd{Verbal}{TS} & Bridge
& \hd{Time}{Weaver} & \hd{Wave}{Stitch}
& & \hd{\textit{Real-}}{\textit{Valid}$^{*}$} \\
\midrule
\multicolumn{10}{l}{\emph{MIMIC-III}: diagnosis and mortality prediction with ABP only} \\
\multirow{2}{*}{Sepsis} & AUROC & 0.566 & 0.513 & 0.515 & 0.505 & 0.510 & 0.525 & \textbf{0.677} & \textit{0.633} \\
                & AUPRC & 0.360 & 0.316 & 0.321 & 0.311 & 0.315 & 0.324 & \textbf{0.494} & \textit{0.436} \\
\multirow{2}{*}{HF}     & AUROC & 0.523 & 0.518 & 0.501 & 0.493 & 0.503 & 0.503 & \textbf{0.650} & \textit{0.650} \\
                & AUPRC & 0.275 & 0.276 & 0.268 & 0.262 & 0.268 & 0.269 & \textbf{0.413} & \textit{0.408} \\
\multirow{2}{*}{Mort.}  & AUROC & 0.519 & 0.501 & 0.511 & 0.515 & 0.497 & 0.553 & \textbf{0.599} & \textit{0.658} \\
                & AUPRC & 0.181 & 0.174 & 0.180 & 0.190 & 0.173 & 0.205 & \textbf{0.249} & \textit{0.316} \\
\midrule
\multicolumn{10}{l}{\emph{MIMIC-IV}: diagnosis and mortality prediction with ABP only} \\
\multirow{2}{*}{Sepsis} & AUROC & 0.531 & 0.520 & 0.527 & 0.498 & 0.504 & 0.495 & \textbf{0.692} & \textit{0.657} \\
                & AUPRC & 0.175 & 0.177 & 0.168 & 0.154 & 0.160 & 0.162 & \textbf{0.286} & \textit{0.260} \\
\multirow{2}{*}{HF}     & AUROC & 0.500 & 0.510 & 0.501 & 0.503 & 0.503 & 0.498 & \textbf{0.637} & \textit{0.596} \\
                & AUPRC & 0.258 & 0.267 & 0.258 & 0.259 & 0.259 & 0.260 & \textbf{0.364} & \textit{0.330} \\
\multirow{2}{*}{Mort.}  & AUROC & 0.504 & 0.560 & 0.506 & 0.484 & 0.519 & 0.535 & \textbf{0.702} & \textit{0.656} \\
                & AUPRC & 0.118 & 0.138 & 0.116 & 0.110 & 0.117 & 0.130 & \textbf{0.236} & \textit{0.209} \\
\midrule
\multicolumn{10}{l}{\emph{AI-READI}: \texttt{study\_group} prediction under 2 input modes} \\
\multirow{2}{*}{\makecell[l]{Study group\\\small(CGM only)}}  & AUROC & 0.502 & 0.519 & 0.546 & 0.715 & 0.656 & 0.604 & \textbf{0.771} & \textit{0.763} \\
                                                     & AUPRC & 0.253 & 0.262 & 0.291 & 0.412 & 0.356 & 0.330 & \textbf{0.504} & \textit{0.477} \\
\multirow{2}{*}{\makecell[l]{Study group\\\small(CGM$+$Lab)}} & AUROC & 0.759 & 0.757 & 0.552 & 0.794 & 0.790 & 0.777 & \textbf{0.795} & \textit{0.796} \\
                                                     & AUPRC & 0.469 & 0.466 & 0.298 & 0.525 & 0.514 & 0.500 & \textbf{0.533} & \textit{0.525} \\
\bottomrule
\end{tabular}
\caption{Downstream evaluation under a train-on-real, test-on-synthetic
protocol: a probe trained on real signals scores each method's generated signal
(5-seed mean). MIMIC-III/-IV predict external ICD diagnoses (sepsis,
heart failure ``HF'') and in-hospital mortality from the generated ABP;
AI-READI predicts four-class study\_group from the generated CGM alone or
with lab test results. Bold marks the best generative method per row.}
\label{tab:downstream-all}
\end{table}

\paragraph{Evaluation protocol}
Because no ground-truth ``clean'' target exists for held-out subjects, we assess
generation quality by downstream clinical utility rather than by point-wise
reconstruction error. Following a train-on-real, test-on-synthetic protocol, a
classifier trained on real signals scores each method's generated
signals against their true labels; every method faces the same probe, so results
are directly comparable, and each metric is the mean over five classifier seeds
(per-seed standard deviations and significance tests are in the supplement).
Because two of the downstream labels (in-hospital mortality on MIMIC and
\texttt{study\_group} on AI-READI) would otherwise be model inputs, we remove the
categorical label from the conditioning set of \emph{every} method and retrain, so
no reported number is scored on a label its generator was given.
We also score the real validation signals under the same probe
(\textit{Real-Valid}$^{*}$) and treat this as an approximate anchor rather than an
upper bound: it is itself a finite-sample estimate, and a generated signal can score
above it. The downstream tasks are four-class \texttt{study\_group} classification on
AI-READI --- from the generated CGM alone and with the real labs fused into the probe
input --- and, on MIMIC-III/-IV, external ICD diagnoses (\emph{sepsis}, \emph{heart
failure}) plus in-hospital \emph{mortality} from the generated ABP alone.

\subsection{Main Results}
\label{sec:main-results}

\ourmethod{} is the strongest generative model in every row of
Table~\ref{tab:downstream-all}, attaining the best AUROC and AUPRC on all sixteen
(dataset, task, metric) combinations. The margin is largest on the critical-care benchmarks, where recovering diagnoses and mortality from the ABP waveform alone is hard and the six baselines hover near chance; on AI-READI our generated CGM
leads clearly without the real labs and matches the strongest baseline once they are
fused in, where every method saturates because the real labs dominate the probe. Against the \textit{Real-Valid}$^{*}$ reference obtained from the real
signal under the same probe, \ourmethod{} reaches or exceeds it on thirteen of the
sixteen rows and falls below it on three, most visibly on MIMIC-III mortality
(AUROC $0.599$ vs.\ $0.658$, AUPRC $0.249$ vs.\ $0.316$).

\begin{figure}[!tbp]
\centering
\begin{minipage}[t]{0.48\textwidth}\vspace{0pt}
\centering
\includegraphics[width=\linewidth]{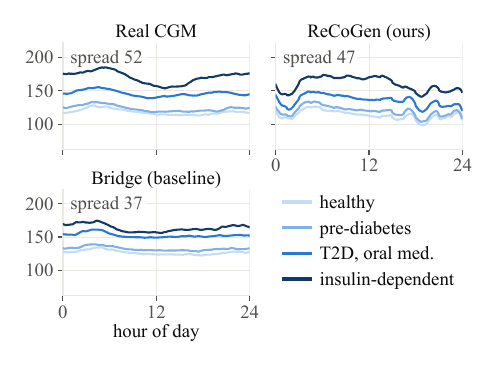}
\caption{Generated CGM reproduces the class structure of the real signal on AI-READI.
Each curve is the 24-hour trace averaged over the held-out participants of one study\_group class; the vertical axis is mean glucose in mg/dL, shared
across panels, and ``spread'' is the gap between the highest and lowest class mean.
Bridge is the strongest baseline on this dataset.}
\label{fig:qualitative-aireadi}
\end{minipage}
\hfill
\begin{minipage}[t]{0.48\textwidth}\vspace{0pt}
\centering
\small
\setlength{\tabcolsep}{5pt}
\renewcommand{\arraystretch}{1.15}
\adjustbox{max width=\linewidth}{\begin{tabular}{ll cccc}
\toprule
Dataset & Task & Full & $-$TS & $-$static & \textit{Real} \\
\midrule
\multirow{3}{*}{MIMIC-III}
& Sepsis & 0.677 & 0.579 & \underline{0.500} & \textit{0.633} \\
& HF     & 0.650 & 0.618 & \underline{0.544} & \textit{0.650} \\
& Mortality  & 0.599 & 0.570 & \underline{0.521} & \textit{0.658} \\
\midrule
\multirow{3}{*}{MIMIC-IV}
& Sepsis & 0.692 & 0.656 & \underline{0.587} & \textit{0.657} \\
& HF     & 0.637 & 0.608 & \underline{0.516} & \textit{0.596} \\
& Mortality  & 0.702 & \underline{0.561} & 0.565 & \textit{0.656} \\
\midrule
AI-READI & Study Group & 0.771 & 0.750 & \underline{0.507} & \textit{0.763} \\
\bottomrule
\end{tabular}}
\captionof{table}{Conditioning ablation: \ourmethod{} retrained with all conditions (Full),
without the time-series conditions ($-$TS), or without the static ones ($-$static),
Each ablated model is retrained and re-scored end to end by the same
train-on-real, test-on-synthetic probe (AUROC, 5-seed mean); Full and Real repeat the
corresponding values of Table~\ref{tab:downstream-all}. Underline marks the more
damaging removal per row.}
\label{tab:ablation-ts-static}
\end{minipage}
\end{figure}

\paragraph{A qualitative look on AI-READI}
Figure~\ref{fig:qualitative-aireadi} is an illustration rather than added evidence,
and it covers one dataset. \texttt{study\_group} is ordinal and the real CGM class
means span $52$ mg/dL, so the class structure is legible in the waveform itself:
\ourmethod{} reproduces the ordering and most of the span ($47$ mg/dL), while Bridge
compresses the classes to $37$ mg/dL and shifts the whole cohort upward. We plot
class means because the variability between participants of one class is far larger
than the separation between classes. The same plot on MIMIC would show little, since
the real ABP level separates the ICU labels by only a few mmHg; there the comparison
rests on Table~\ref{tab:downstream-all}, and we make no visual claim.

\paragraph{What the protocol does and does not establish}
The probe scores a generated signal, and that signal is a function of the conditions,
so a high score can arise in two ways: the synthesized target carries the same
label-relevant physiology as the real one, or the generator re-expresses information
already in the conditions in a form the probe reads easily. Downstream utility
therefore measures label-relevant information transfer, and conflates target fidelity
with condition re-expression. We add no pointwise fidelity metric because the setting
offers no clean target: the signal we synthesize for a held-out subject is precisely
the one that was not measured. Two observations bound the ambiguity. Sepsis and heart
failure are external ICD diagnoses never available to the generator, so recovering
them from the generated waveform cannot be explained by \emph{copying} a conditioning
variable; they can still be correlated with the covariates it does see. Mortality is
the sharper case: driven by the hemodynamic state ABP encodes and correlated with
those covariates, it is the label most exposed to a conditioning-correlated shortcut.
Table~\ref{tab:ablation-ts-static} bounds that room: static conditions alone already
support much of the score (MIMIC-IV sepsis AUROC $0.656$ with no time-series
condition). Scores above \textit{Real-Valid}$^{*}$ are thus evidence that
label-relevant structure survives generation, possibly in a cleaner form than the
measured waveform, not that the synthetic signal is the more faithful one.

\subsection{Ablation Study}

\begin{figure}[!t]
\centering
\includegraphics[width=\linewidth]{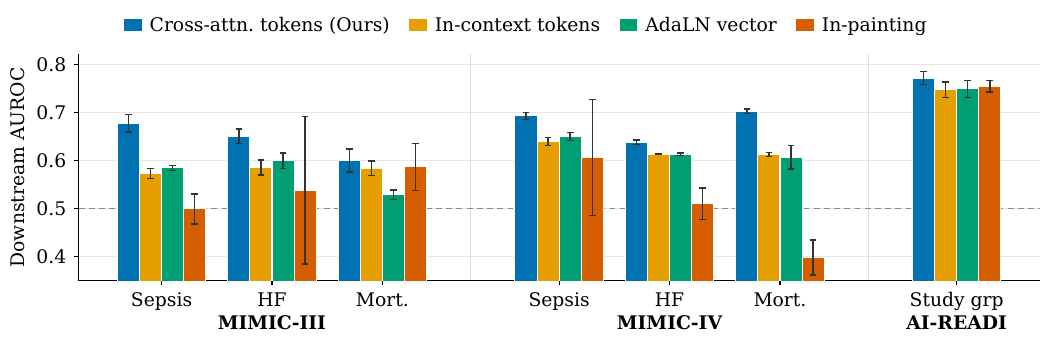}
\caption{Time-series conditioning mechanism: per-task downstream AUROC (bars: 5-seed
mean; error bars: $\pm1$ std) for four injections differing only in how the
conditioning series enters --- frozen-AE latents via learnable cross-attention tokens
(\ourmethod{}), the same latents as in-context tokens, the pooled latents as one
AdaLN vector, and RePaint-style in-painting (no AE). Dashed line: chance.}
\label{fig:ablation-tscond}
\end{figure}

\paragraph{Time-series vs.\ static conditioning}
To quantify how much each conditioning group contributes, we retrain \ourmethod{}
leave-one-group-out (Table~\ref{tab:ablation-ts-static}): from the ``Full''
conditioning we drop either all time-series conditions ($-$TS) or all static/tabular
ones ($-$static), applying the ablation identically at training, validation and
generation time. Removing either group lowers AUROC on every task, so both carry
information that \ourmethod{} transfers into its generation; dropping the static
conditions hurts more in most cases.

\begin{table}[!tbp]
\centering
\begin{minipage}[t]{0.48\textwidth}\vspace{0pt}
\centering
\small
\setlength{\tabcolsep}{5pt}
\renewcommand{\arraystretch}{1.15}
\adjustbox{max width=\linewidth}{\begin{tabular}{ll ccc}
\toprule
Dataset & Task & \makecell{Token$+$AdaLN\\(\ourmethod{})} & \makecell{Token\\-only} & \makecell{AdaLN\\-only} \\
\midrule
\multirow{3}{*}{MIMIC-III}
& Sepsis    & \textbf{0.677} & 0.547 & 0.595 \\
& HF        & \textbf{0.650} & 0.536 & 0.648 \\
& Mortality & \textbf{0.599} & 0.525 & 0.561 \\
\midrule
\multirow{3}{*}{MIMIC-IV}
& Sepsis    & \textbf{0.692} & 0.560 & 0.562 \\
& HF        & \textbf{0.637} & 0.551 & 0.593 \\
& Mortality & \textbf{0.702} & 0.501 & 0.509 \\
\midrule
AI-READI & Study group & \textbf{0.771} & 0.624 & 0.742 \\
\bottomrule
\end{tabular}}
\caption{Static-feature encoding ablation as a token$\times$AdaLN factorial,
per-task downstream AUROC: in-context tokens$+$AdaLN context
(\ourmethod{}) vs.\ tokens-only vs.\ AdaLN-only.
Bold marks the best injection per row.}
\label{tab:ablation-static}
\end{minipage}
\hfill
\begin{minipage}[t]{0.48\textwidth}\vspace{0pt}
\centering
\includegraphics[width=\linewidth]{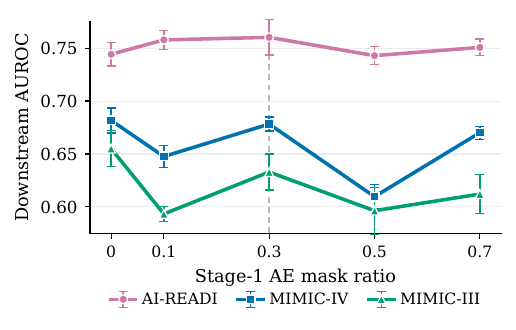}
\captionof{figure}{Stage-I mask-ratio sensitivity. Downstream AUROC against the stage-I mask
ratio; each line is the mean over a dataset's tasks (AI-READI: study\_group;
MIMIC-III/-IV: sepsis, HF, mortality), error bars the pooled $5$-seed probe std. The
sweep is a single end-to-end re-run, so its $\rho{=}0.3$ point need not reproduce the
main table exactly.}
\label{fig:ablation-maskratio}
\end{minipage}
\end{table}

\paragraph{Time-series conditioning mechanism}
Holding everything else fixed, we compare four ways for the conditioning series to
enter (Figure~\ref{fig:ablation-tscond}): frozen-AE latents read by learnable
per-modality queries via \emph{cross-attention} into in-context tokens
(\ourmethod{}); the same latents as plain in-context tokens; the pooled latents
through AdaLN modulation; and the in-painting alternative of Diffusion-TS
\citep{yuan2024diffusionts} and ImagenTime \citep{naiman2024imagentime}.
Cross-attention wins overall, most clearly on the ABP tasks, where the other
injections fall well behind and in-painting is unstable, dropping toward or below
chance on the hardest cases (\textit{e.g.}\ MIMIC-IV mortality). On AI-READI all
four are comparable.

\paragraph{Static-feature encoding}
We inject the static features as a token$\times$AdaLN factorial
(Table~\ref{tab:ablation-static}, same protocol): in-context tokens plus a pooled
AdaLN modulation (\ourmethod{}), tokens only, or the AdaLN modulation alone. Both
paths together win on every (dataset, task), by a wide margin on MIMIC-IV, so they
are complementary rather than redundant; between the single-path variants, removing
the AdaLN modulation is consistently more damaging. The static features are therefore
carried primarily through AdaLN, the token adding a complementary gain only when that
path is present.

\paragraph{Stage-I mask-ratio sensitivity}
The stage-I autoencoders hold out and reconstruct an additional fraction of the
observed points, the \emph{mask ratio}, set to $0.3$ in \ourmethod{}. Sweeping it
over $\{0,0.1,0.3,0.5,0.7\}$ and retraining both stages end to end per value
(Figure~\ref{fig:ablation-maskratio}, AUROC averaged over each dataset's tasks), the
generator is fairly robust: AUROC varies within a modest range with no monotone
trend, and $0.3$ is a consistently strong setting.


\section{Related Works}

\paragraph{Conditional Time-Series Generation}
We review existing generators through the lens of how they ingest conditioning signals.
\textbf{1)} Label-conditioned models steer generation with a discrete class: TTS-CGAN~\cite{li2022ttscgan} and TimeVQVAE~\cite{lee2023timevqvae} fall into this category.
\textbf{2)} Attribute-conditioned models condition on structured metadata: TEdit \cite{jing2024tedit} and WaveStitch \cite{shankar2025wavestitch} are two representative works. TimeWeaver \cite{narasimhan2024timeweaver} extends this line of work by enabling multimodal conditioning through the tokenization and fusion of heterogeneous covariates spanning categorical, continuous, and time-variant modalities.
\textbf{3)} Text-conditioned models verbalize the condition into natural language and inject it through text embedding models: VerbalTS~\cite{gu2025verbalts}, T2S~\cite{ge2025t2s}, Bridge~\cite{li2025bridge}, and DiffuSETS~\cite{lai2025diffusets} are in this group.
\textbf{4)} Signal-conditioned models take the observed portion of a time series as the condition, prediction and imputation being the representative tasks: CSDI~\cite{tashiro2021csdi}, Diffusion-TS~\cite{yuan2024diffusionts}, and ImagenTime~\cite{naiman2024imagentime}.
Across all four groups, a generator is tailored to \emph{one} kind of condition and
absorbs it directly into the generative process, either as an in-context sequence to
be in-painted alongside the target or as fused metadata; even
TimeWeaver~\cite{narasimhan2024timeweaver}, which admits multiple covariate types,
assumes them densely aligned and does not model the heavy, modality-dependent
missingness of co-recorded physiological streams. \ourmethod{} departs from this
recipe in two ways: it targets genuinely multimodal conditioning (several irregularly
sampled, partially observed time series together with static clinical descriptors)
rather than a single condition type, and it decouples \emph{how a condition is
represented} from \emph{how the target is generated}, freezing a missingness-aware
encoder per modality so the generator consumes robust condition tokens instead of
reconstructing the raw streams itself.

\paragraph{Medical Downstream Tasks}
Our probe is trained on real signals, which is meaningful only if the labels are ones those signals genuinely predict; we therefore ground each task in prior work.
CGM traces routinely characterize glycemic state and screen for diabetes \citep{hall2018glucotypes}, and CGMformer \citep{lu2025cgmformer} fine-tunes exactly the diabetes-status target (study\_group) we probe on AI-READI.
ABP and its co-recorded vitals are the core input to ICU risk prediction: MIMIC benchmarks predict in-hospital mortality from clinical time series \citep{harutyunyan2019multitask}, \citet{nemati2018sepsis} predict sepsis onset from routine vitals, and others predict circulatory failure and hypotension from the same signals~\citep{hyland2020circulatory,hatib2018hypotension}.
Our three ABP labels are complementary: mortality is the canonical ICU outcome, directly driven by the hemodynamic state ABP encodes \citep{vincent2013shock,maheshwari2018hypotension,alghatani2021los}, while sepsis and heart failure are external diagnoses never given to the generator, so they test whether it preserves label-relevant physiology rather than low-order statistics.

\section{Conclusion}
\label{sec:conclusion}
\ourmethod{} addresses conditional generation of physiological time series from
several irregularly sampled, partially observed signals together with static clinical
variables. Feeding them straight into the generator makes one objective both
represent the conditions and generate the target, serving neither well;
\ourmethod{} separates the two, distilling each irregular stream into
missingness-tolerant tokens with a per-modality masked autoencoder and generating the
target with a frozen-encoder flow-matching model. It attains the best downstream
utility on all sixteen (dataset, task, metric) settings across AI-READI, MIMIC-III
and MIMIC-IV, by the widest margin on the critical-care ABP benchmarks where the six
baselines hover near chance. Two limitations frame that result: downstream utility
conflates target fidelity with a re-expression of the conditions, so the real-signal
reference is an anchor rather than a ceiling; and encoding each modality
independently leaves cross-modal dependencies to the generator. Modeling the
conditions jointly, and separating fidelity from re-expression in a paired setting,
are natural next steps.

\section*{Acknowledgment}

This research was partially funded by the National Institutes of Health (NIH) under award 1R01EB037101-01. The views and conclusions contained in this document are those of the authors and should not be
interpreted as representing the official policies, either expressed or implied, of the NIH.

\bibliography{999_reference}
\bibliographystyle{style/icml2025}

\titlespacing*{\section}{0pt}{*1}{*1}
\titlespacing*{\subsection}{0pt}{*1.25}{*1.25}
\titlespacing*{\subsubsection}{0pt}{*1.5}{*1.5}

\setlength{\abovedisplayskip}{\baselineskip} 
\setlength{\abovedisplayshortskip}{0.5\baselineskip} 
\setlength{\belowdisplayskip}{\baselineskip}
\setlength{\belowdisplayshortskip}{0.5\baselineskip}

\clearpage
\appendix
\label{sec:append}
\part*{Appendix}
{
\setlength{\parskip}{-0em}
\startcontents[sections]
\printcontents[sections]{ }{1}{}
}

\setlength{\parskip}{.35em}

This supplement lists the full parameter settings behind every number reported in
the main paper: the windowing and preprocessing of the three cohorts, the Stage~I
masked autoencoders, the Stage~II conditional flow-matching generator, the sampling
configuration, the six baselines, and the downstream evaluation probe. Values that a
launch script leaves unset are the corresponding reference implementation's
defaults. All runs use a single NVIDIA GPU per job.

\section{Data Windowing and Preprocessing}

Every method in a given benchmark consumes exactly the same windows: the baselines
call the dataset builder, split, windowing and valid-window filter of \ourmethod{}
verbatim, and generation is driven by the same window index (the metadata file
written by our own generation pass), so the downstream probe scores all methods on
an identical set of subjects and time intervals. Each channel is min-max
normalized to $[-1,1]$ per series, and each per-modality pack carries a binary
observed mask ($1$ = observed, $0$ = missing or padding).

\paragraph{AI-READI (CGM generation)}
Windows are built on the glucose anchor grid: anchor sampling period $5$~minutes
with a $2$~s matching tolerance, maximum anchor gap $10$~minutes, maximum window
span $24$~hours, and at least $288$ events per day, giving $T{=}288$ steps per
window. Windows whose target exceeds a $0.5$ missing ratio are discarded. The four
conditioning modalities (heart rate, calorie expenditure, physical activity,
respiratory rate) are aligned to that grid; the static descriptor consists of $6$
numeric CGM-enhanced features, $6$ self-report binary flags (yes/no/unknown), and
$6$ ``past two weeks'' medication labels. The $4$-class \texttt{study\_group} is the
downstream target, not a condition (see ``Label conditioning is disabled'' below).

\paragraph{MIMIC-III and MIMIC-IV (ABP generation)}
Stays are split $80/20$ into train and test by \texttt{train\_frac}${=}0.8$; a
window is kept if at least $5\%$ of its target steps are observed
(\texttt{min\_target\_coverage}${=}0.05$) and its missing ratio does not exceed
$0.5$. The target is the mean ABP channel on a $24$-hour, $5$-minute grid
($T{=}288$). On MIMIC-III the ABP and the three vitals (heart rate, respiratory
rate, $\mathrm{SpO}_2$) come from the Waveform Database Matched Subset numerics
resampled onto that grid; on MIMIC-IV they come from bedside charted measurements,
with the vitals kept on the coarser hourly grid
($T_{\mathrm{cond}}{=}24$). The static descriptor is a $27$-dimensional vector (age,
gender, and $25$ chart/lab/drug window means); the binary in-hospital-mortality label
is likewise a downstream target, not a condition.

\section{Stage~I: Per-Modality Masked Autoencoders}

One autoencoder is trained per time-variant conditioning modality (four on
AI-READI, three on each MIMIC cohort) with the settings in
Table~\ref{tab:supp-stage1}. The network projects the (single-channel) input to
$d_{\text{model}}$, adds sinusoidal positional encodings, replaces every naturally
missing or additionally held-out step with one shared learnable
\texttt{[missing]} token, applies a Transformer encoder, and reads out with a linear
head. The loss is the mean squared error on the additionally held-out steps only;
the mask ratio $\rho$ is the fraction of \emph{observed} steps that are held out.
The checkpoint with the lowest validation loss is kept and frozen for Stage~II.

\begin{table}[!tbp]
\centering
\begin{minipage}[t]{0.48\textwidth}\vspace{0pt}
\centering
\small
\setlength{\tabcolsep}{5pt}
\adjustbox{max width=\linewidth}{\begin{tabular}{lcc}
\toprule
Setting & AI-READI & MIMIC \\
\midrule
Model dimension $d_{\text{model}}$   & 128 & 128 \\
Attention heads                      & 4   & 4 \\
Encoder layers                       & 3   & 3 \\
Input series length $L$              & 288 & 288 / 24 \\
Positional-encoding capacity         & 288 & 512 \\
Mask ratio $\rho$                    & 0.3 & 0.3 \\
Epochs                               & 1000 & 1000 \\
Batch size                           & 256 & 256 \\
Optimizer                            & AdamW & AdamW \\
Learning rate                        & $3{\times}10^{-4}$ & $3{\times}10^{-4}$ \\
Weight decay                         & $10^{-4}$ & $10^{-4}$ \\
Max.\ missing ratio (filter)         & 0.5 & 0.1 \\
Seed                                 & 0 & 0 \\
\bottomrule
\end{tabular}}
\caption{Stage~I masked-autoencoder settings. One model is trained per
conditioning modality; the mask-ratio sweep of the main paper retrains all of them
at $\rho\in\{0,0.1,0.3,0.5,0.7\}$. The conditioning series are $L{=}288$ steps long
on MIMIC-III (5-minute grid) and $L{=}24$ on MIMIC-IV (hourly grid); the
positional-encoding table is simply sized to cover them.}
\label{tab:supp-stage1}
\end{minipage}
\hfill
\begin{minipage}[t]{0.48\textwidth}\vspace{0pt}
\centering
\small
\setlength{\tabcolsep}{3pt}
\adjustbox{max width=\linewidth}{\begin{tabular}{lcc}
\toprule
Setting & AI-READI & MIMIC \\
\midrule
\multicolumn{3}{l}{\emph{Backbone (\texttt{JiT})}} \\
Image size (delay embedding)         & 18 & 18 \\
Patch size                           & 2 & 2 \\
Hidden size                          & 128 & 128 \\
Depth                                & 4 & 4 \\
Attention heads                      & 4 & 4 \\
Attn./proj.\ dropout                 & 0.1 / 0.1 & 0.1 / 0.1 \\
Bottleneck dimension                 & 64 & 64 \\
Label classes                        & 4 & 2 \\
In-context tokens                    & 18 & 13 \\
\midrule
\multicolumn{3}{l}{\emph{Condition encoder}} \\
Tokens per modality                  & 4 & 4 \\
Encoder width / heads                & 128 / 4 & 128 / 4 \\
Frozen AE (dim/heads/layers)         & 128/4/3 & 128/4/3 \\
\midrule
\multicolumn{3}{l}{\emph{Flow matching and optimization}} \\
$P_{\text{mean}}$ / $P_{\text{std}}$   & 0.5 / 1.2 & 0 / 1.2 \\
$t$ clamp $\epsilon_t$               & $10^{-5}$ & $10^{-5}$ \\
Noise scale                          & 1.0 & 1.0 \\
Label drop prob.                     & 0.0 & 0.0 \\
EMA decays                           & \multicolumn{2}{c}{0.999 and 0.9999} \\
Epochs                               & 1000 & 1000 \\
Batch size                           & 256 & 256 \\
Optimizer                            & AdamW & AdamW \\
Learning rate                        & $10^{-4}$ & $10^{-4}$ \\
Weight decay                         & $10^{-5}$ & $10^{-5}$ \\
Gradient clipping                    & 1.0 & 1.0 \\
Seed                                 & 42 & 42 \\
\midrule
\multicolumn{3}{l}{\emph{Sampling}} \\
ODE solver                           & Heun & Heun \\
Sampling steps                       & 50 & 50 \\
Guidance scale                       & 1.0 & 1.0 \\
$t$ interval                         & $[0,1]$ & $[0,1]$ \\
Weights / epoch                      & \multicolumn{2}{c}{EMA($0.999$), epoch 1000} \\
\bottomrule
\end{tabular}}
\caption{Stage~II generator settings. $P_{\text{mean}}$ is the only value that
differs between cohorts; it was selected on validation utility in a per-cohort
sweep, with everything else held fixed.}
\label{tab:supp-stage2}
\end{minipage}
\end{table}

\section{Stage~II: Conditional Flow-Matching Generator}

Stage~II freezes the Stage~I encoders and trains the \texttt{JiT} denoiser with the
settings in Table~\ref{tab:supp-stage2}. A $T{=}288$ series is delay-embedded into
an $18\times18$ single-channel image (non-overlapping windows, i.e.\ delay${=}$
embedding${=}$image size) and the flow-matching loss is computed only inside the
valid-region mask. Each conditioning modality contributes $4$ in-context tokens
produced by learnable per-modality queries cross-attending over its frozen latents;
the static descriptor contributes $2$ further tokens on AI-READI (one numeric MLP
token and one categorical-transformer token) and $1$ on the MIMIC cohorts, and is
additionally pooled into the AdaLN modulation. The in-context stream therefore has
length $4{\times}4{+}2{=}18$ on AI-READI and $3{\times}4{+}1{=}13$ on MIMIC, and is
inserted before the first transformer block (\texttt{in\_context\_start}${=}0$).
Two exponential moving averages of the weights are tracked; all reported numbers
use the faster one (decay $0.999$) at epoch $1000$.

\paragraph{Label conditioning is disabled}
Both downstream label sets (in-hospital mortality on the MIMIC cohorts, the $4$-class
\texttt{study\_group} on AI-READI) would leak if they also conditioned the generator,
so the categorical label is removed from \emph{every} method and all label-conditioned
models are retrained from scratch. For \ourmethod{} this means the label embedding is
pinned to its null slot identically at training, validation and generation time, so
the true label never reaches the network; the ``Label classes'' row of
Table~\ref{tab:supp-stage2} therefore sizes an embedding that is never driven by a
real label. Among the baselines, TimeWeaver's categorical metadata attribute is held
constant, the verbalization fed to VerbalTS and Bridge omits the outcome clause, and
Diffusion-TS, ImagenTime and WaveStitch never had a label path to begin with. Every
number reported in the main paper comes from this de-leaked setting.

\paragraph{Ablation variants}
All ablations reuse Table~\ref{tab:supp-stage2} and change exactly one component.
The conditioning ablations ($-$TS, $-$static) replace a group with the model's
built-in missing representation (zeroed values and masks, or the null label slot),
so tensor shapes and the in-context length stay fixed, and the ablation is applied
identically at training, validation and generation time. The injection variants
(in-context tokens without cross-attention, AdaLN-only pooling, RePaint-style
in-painting) and the static-encoding factorial (Token$+$AdaLN, Token-only,
AdaLN-only) likewise keep every other setting. Each variant is retrained from
scratch; the mask-ratio sweep retrains both stages end to end per value.

\section{Baselines}

\begin{wraptable}{r}{\aaaicolwidth}
\centering
\small
\setlength{\tabcolsep}{5pt}
\begin{tabular}{ll}
\toprule
Method & Settings \\
\midrule
Diffusion-TS & $d_{\text{model}}{=}128$; $3$ encoder / $4$ decoder layers; \\
             & $1000$ diffusion steps, $200$ sampling steps \\
ImagenTime   & delay${=}$embedding${=}32$ ($32\times32$ image); \\
             & UNet width $64$; $100$ diffusion steps \\
TimeWeaver   & channels $64$, $4$ layers, $8$ heads; cosine schedule; \\
             & $1000$ diffusion steps; metadata: $\text{cond}{=}128$, \\
             & $\text{token}{=}64$, $2$ fusion layers, $4$ fusion heads \\
WaveStitch   & SSSD-S4: residual/skip channels $64$, $4$ residual \\
             & layers, state size $64$, cond.\ channels $16$; \\
             & $200$ diffusion steps \\
VerbalTS     & channels $64$, $4$ layers, AdaLN text conditioning; \\
             & cosine schedule, $1000$ diffusion steps \\
Bridge       & UNet width $64$, $2$ residual blocks per stage; \\
             & cosine schedule, $1000$ diffusion steps; prototype \\
             & series drawn from the training split \\
\bottomrule
\end{tabular}
\caption{Baseline settings shared by the three benchmarks. Unlisted values are the
reference implementation's defaults.}
\label{tab:supp-baselines}
\end{wraptable}

All six baselines are trained for $1000$ epochs at learning rate $10^{-4}$ on the
same windows, with the seed fixed to $42$ and the target reduced to the single
generated channel. Batch size is $512$ for Diffusion-TS, ImagenTime and TimeWeaver
and $256$ for WaveStitch, VerbalTS and Bridge. Five of the six use
Adam ($\beta{=}(0.9,0.96)$); ImagenTime keeps AdamW, as in its reference
implementation. Method-specific settings are listed in
Table~\ref{tab:supp-baselines}; for the text-conditioned pair, the verbalized
condition is embedded by a frozen LongCLIP text tower ($768$-d, $248$-token window)
in which each conditioning series is rendered as summary statistics plus a
$20$-point downsampled numeric transcript in physical units.

\section{Downstream Evaluation Protocol}

The probe is a 1D CNN: three convolutional blocks with kernel sizes $7$, $5$ and
$3$ and $64$, $128$ and $256$ channels, each with batch normalization and ReLU and
the first two followed by max pooling, then global average pooling, dropout $0.2$
and a linear classifier. It is trained on \emph{real} signals for $50$ epochs with
Adam, learning rate $10^{-3}$, weight decay $10^{-4}$ and batch size $256$, and
then applied to each method's \emph{generated} signals. Every reported metric is the
mean (and standard deviation) over the five probe seeds $\{0,1,2,3,4\}$; the same
trained probe configuration scores every method, and the real validation signals
scored under it give the \textit{Real-Valid}$^{*}$ reference. In the fused AI-READI
mode the generated CGM passes through the same convolutional trunk while the
biochemical lab values and their missingness mask pass through a two-layer MLP
(width $64$, dropout $0.2$); the two embeddings are concatenated before the linear
classifier head. The MIMIC cohorts are reported in the ABP-only mode.

\section{Statistical Significance of the Main Results}

The main paper reports 5-seed means. Table~\ref{tab:supp-sig} adds the per-seed
dispersion behind every AUROC entry of that table and tests \ourmethod{}'s margin
over each baseline; Table~\ref{tab:supp-strongest} isolates the comparison against
the \emph{strongest} rival in each setting, and Table~\ref{tab:supp-wilcoxon}
aggregates across settings.

\paragraph{Test protocol}
The unit of replication is the downstream probe seed: all methods are scored by the
same probe configuration re-trained under the five seeds $\{0,1,2,3,4\}$, giving
$n{=}5$ AUROC values per (dataset, task, method). For each of the eight (dataset,
task) settings we compare \ourmethod{} against each of the six baselines with a
two-sided Welch $t$-test (unequal variances, Satterthwaite degrees of freedom) and
control the family-wise error rate across the six comparisons within that setting by
the Holm--Bonferroni procedure; $^{\ast}$, $^{\ast\ast}$ and $^{\ast\ast\ast}$ in
Table~\ref{tab:supp-sig} denote Holm-adjusted $p<0.05$, $p<0.01$ and $p<0.001$. To
summarize across settings without assuming normality, we also run an exact Wilcoxon
signed-rank test on the eight paired per-setting AUROC differences
(Table~\ref{tab:supp-wilcoxon}); with $n{=}8$ its smallest attainable two-sided
$p$-value is $2/2^{8}=0.0078$, which every baseline attains.

\paragraph{Result}
\ourmethod{}'s margin is significant after Holm correction in 46 of the 48
per-setting comparisons. The two exceptions are both on AI-READI in the
CGM$+$Lab mode, against Bridge ($+0.001$, $p{=}0.69$) and TimeWeaver ($+0.005$,
$p{=}0.38$): once the real laboratory values are fused into the probe input every
method saturates near the real-signal reference, so that mode does not separate
generators. In the CGM-only mode on the same cohort the margin over the strongest
baseline is $+0.056$ and significant. The single narrowest \emph{significant} margin
is MIMIC-III mortality against WaveStitch ($+0.046$, $p{=}0.009$), which is also the
setting where \ourmethod{} falls furthest below the real-signal reference. Across
settings, \ourmethod{} beats every baseline in $8/8$ settings with median margins of
$+0.137$ to $+0.165$ AUROC.

\paragraph{What these tests do and do not cover}
Five limitations should be read alongside the numbers. \textbf{(i)} Only AUROC is
tested; the per-seed dispersion of AUPRC was not retained by the evaluation
pipeline, so the AUPRC rows of the main table carry no test. \textbf{(ii)} The five
probe seeds are shared across methods, so the observations are in fact paired; we
nevertheless use the unpaired Welch test, which discards that pairing and is
therefore the conservative choice --- a paired test would yield smaller $p$-values.
\textbf{(iii)} The tests are computed from means and standard deviations stored to
three decimal places, so for the lowest-variance entries (subscript $.001$) the $t$
statistics should be read as orders of magnitude rather than exact values; none of
the Holm-adjusted conclusions turns on those rows. \textbf{(iv)} On AI-READI,
\ourmethod{} was scored by its own five-seed probe instance rather than the one used
for the baselines; the two agree on the real validation signal to within $0.005$
AUROC ($0.766$ vs.\ $0.763$ in CGM-only, $0.798$ vs.\ $0.796$ in CGM$+$Lab), and the
main paper quotes the baseline probe's value as the reference, but the AI-READI
comparisons are across probe draws rather than within one. \textbf{(v)} The
\textit{Real-Valid}$^{*}$ reference is not tested: its per-seed dispersion was not
retained either, and, as the main paper argues, it is an approximate anchor rather
than a null hypothesis worth rejecting.

\begin{table}[!tb]
\centering
\small
\setlength{\tabcolsep}{4pt}
\begin{tabular}{ll cccccc c c}
\toprule
& & \multicolumn{6}{c}{Generative baselines} & & Ref. \\
\cmidrule(lr){3-8}
Dataset & Task & \makecell{Diffusion\\-TS} & \makecell{Imagen\\Time} & \makecell{Verbal\\TS} & Bridge & \makecell{Time\\Weaver} & \makecell{Wave\\Stitch}
& \textbf{\ourmethod{}} & \makecell{\textit{Real-}\\\textit{Valid}$^{*}$} \\
\midrule
MIMIC-III & Sepsis & $.566^{\ast\ast\ast}_{.010}$ & $.513^{\ast\ast\ast}_{.008}$ & $.515^{\ast\ast\ast}_{.004}$ & $.505^{\ast\ast\ast}_{.003}$ & $.510^{\ast\ast\ast}_{.003}$ & $.525^{\ast\ast\ast}_{.005}$ & $\mathbf{.677}_{.019}$ & \textit{.633} \\
 & HF & $.523^{\ast\ast\ast}_{.002}$ & $.518^{\ast\ast\ast}_{.004}$ & $.501^{\ast\ast\ast}_{.003}$ & $.493^{\ast\ast\ast}_{.002}$ & $.503^{\ast\ast\ast}_{.003}$ & $.503^{\ast\ast\ast}_{.002}$ & $\mathbf{.650}_{.015}$ & \textit{.650} \\
 & Mortality & $.519^{\ast\ast}_{.008}$ & $.501^{\ast\ast}_{.006}$ & $.511^{\ast\ast}_{.006}$ & $.515^{\ast\ast}_{.003}$ & $.497^{\ast\ast}_{.004}$ & $.553^{\ast\ast}_{.012}$ & $\mathbf{.599}_{.024}$ & \textit{.658} \\
\midrule
MIMIC-IV & Sepsis & $.531^{\ast\ast\ast}_{.006}$ & $.520^{\ast\ast\ast}_{.004}$ & $.527^{\ast\ast\ast}_{.013}$ & $.498^{\ast\ast\ast}_{.002}$ & $.504^{\ast\ast\ast}_{.007}$ & $.495^{\ast\ast\ast}_{.013}$ & $\mathbf{.692}_{.007}$ & \textit{.657} \\
 & HF & $.500^{\ast\ast\ast}_{.002}$ & $.510^{\ast\ast\ast}_{.001}$ & $.501^{\ast\ast\ast}_{.001}$ & $.503^{\ast\ast\ast}_{.001}$ & $.503^{\ast\ast\ast}_{.002}$ & $.498^{\ast\ast\ast}_{.007}$ & $\mathbf{.637}_{.005}$ & \textit{.596} \\
 & Mortality & $.504^{\ast\ast\ast}_{.016}$ & $.560^{\ast\ast\ast}_{.007}$ & $.506^{\ast\ast\ast}_{.002}$ & $.484^{\ast\ast\ast}_{.003}$ & $.519^{\ast\ast\ast}_{.012}$ & $.535^{\ast\ast\ast}_{.017}$ & $\mathbf{.702}_{.005}$ & \textit{.656} \\
\midrule
AI-READI & SG (CGM) & $.502^{\ast\ast\ast}_{.002}$ & $.519^{\ast\ast\ast}_{.003}$ & $.546^{\ast\ast\ast}_{.008}$ & $.715^{\ast\ast\ast}_{.003}$ & $.656^{\ast\ast\ast}_{.008}$ & $.604^{\ast\ast\ast}_{.015}$ & $\mathbf{.771}_{.013}$ & \textit{.763} \\
 & SG (CGM$+$Lab) & $.759^{\ast\ast\ast}_{.006}$ & $.757^{\ast\ast\ast}_{.005}$ & $.552^{\ast\ast\ast}_{.012}$ & $.794^{\phantom{\ast}}_{.005}$ & $.790^{\phantom{\ast}}_{.007}$ & $.777^{\ast}_{.010}$ & $\mathbf{.795}_{.002}$ & \textit{.796} \\
\bottomrule
\end{tabular}
\caption{Downstream AUROC over the five probe seeds, written
$\text{mean}_{\text{std}}$ with leading zeros omitted; the means are exactly the
AUROC rows of the main downstream table.
Superscripts on a baseline mark the Holm-adjusted significance of \ourmethod{}'s
margin over that baseline within its row (two-sided Welch $t$-test, $n{=}5$ per
method, six comparisons per row): $^{\ast}p<0.05$, $^{\ast\ast}p<0.01$,
$^{\ast\ast\ast}p<0.001$; no superscript means not significant at $0.05$. The
\textit{Real-Valid}$^{*}$ reference is a single value per row --- its per-seed
dispersion was not retained --- and is not tested.}
\label{tab:supp-sig}
\end{table}

\begin{table}[!tbp]
\centering
\begin{minipage}[t]{0.48\textwidth}\vspace{0pt}
\centering
\small
\setlength{\tabcolsep}{3pt}
\adjustbox{max width=\linewidth}{\begin{tabular}{l l r r}
\toprule
Setting & Rival & $\Delta$ & $p_{\mathrm{Holm}}$ \\
\midrule
MIMIC-III Sepsis & Diffusion-TS & $+0.111$ & $<10^{-4}$ \\
MIMIC-III HF & Diffusion-TS & $+0.127$ & $<10^{-5}$ \\
MIMIC-III Mort. & WaveStitch & $+0.046$ & $0.009$ \\
\midrule
MIMIC-IV Sepsis & VerbalTS & $+0.165$ & $<10^{-7}$ \\
MIMIC-IV HF & ImagenTime & $+0.127$ & $<10^{-7}$ \\
MIMIC-IV Mort. & WaveStitch & $+0.167$ & $<10^{-6}$ \\
\midrule
AI-READI SG & Bridge & $+0.056$ & $<10^{-4}$ \\
AI-READI SG$+$Lab & Bridge & $+0.001$ & $0.694$ \\
\bottomrule
\end{tabular}}
\caption{\ourmethod{} against the hardest baseline in each setting; $\Delta$ is the
AUROC margin. ``Rival'' is the baseline with the largest Holm-adjusted $p$-value in
that row of Table~\ref{tab:supp-sig}, i.e.\ the one whose margin is hardest to
establish. Only the saturated AI-READI CGM$+$Lab mode fails to separate.}
\label{tab:supp-strongest}
\end{minipage}
\hfill
\begin{minipage}[t]{0.48\textwidth}\vspace{0pt}
\centering
\small
\setlength{\tabcolsep}{5pt}
\adjustbox{max width=\linewidth}{\begin{tabular}{l c r r r}
\toprule
Baseline & Wins & Median $\Delta$ & $W$ & $p$ \\
\midrule
Diffusion-TS & 8/8 & $+0.137$ & $0$ & $0.0078$ \\
ImagenTime & 8/8 & $+0.142$ & $0$ & $0.0078$ \\
VerbalTS & 8/8 & $+0.165$ & $0$ & $0.0078$ \\
Bridge & 8/8 & $+0.157$ & $0$ & $0.0078$ \\
TimeWeaver & 8/8 & $+0.147$ & $0$ & $0.0078$ \\
WaveStitch & 8/8 & $+0.152$ & $0$ & $0.0078$ \\
\bottomrule
\end{tabular}}
\caption{Exact two-sided Wilcoxon signed-rank test over the eight (dataset, task)
settings, pairing \ourmethod{}'s AUROC with each baseline's. ``Wins'' counts
settings with a positive difference. $p{=}0.0078$ is the smallest value attainable
at $n{=}8$, so it is reached whenever a method wins every setting; the test
establishes consistency across settings, not the size of any single margin.}
\label{tab:supp-wilcoxon}
\end{minipage}
\end{table}

\end{document}